\pdfoutput=1

\documentclass[11pt]{article}

\usepackage{acl}

\usepackage{times}
\usepackage{latexsym}

\usepackage[T1]{fontenc}

\usepackage[utf8]{inputenc}

\usepackage{microtype}

\usepackage{paralist}
\usepackage{enumitem}

\usepackage[ruled]{algorithm2e}
\usepackage{amsmath}
\usepackage{booktabs}
\usepackage{graphicx}
\usepackage{listings}
\usepackage{longtable}
\usepackage{makecell}
\usepackage{multirow}
\usepackage{natbib}
\usepackage{pifont}
\usepackage{xurl}

\title{Debatrix: Multi-dimensional Debate Judge with Iterative Chronological Analysis Based on LLM}

\author{
Jingcong Liang\textsuperscript{\rm 1},
~Rong Ye\textsuperscript{\rm 1,3},
~Meng Han\textsuperscript{\rm 2},
~Ruofei Lai\textsuperscript{\rm 2},\\
\bf{
Xinyu Zhang\textsuperscript{\rm 2},
~Xuanjing Huang\textsuperscript{\rm 1},
~Zhongyu Wei\textsuperscript{\rm 1}\thanks{\enspace Corresponding author.}
}\\
\textsuperscript{1}Fudan University,
~\textsuperscript{2}Huawei Poisson Lab,
~\textsuperscript{3}ByteDance\\
\texttt{jcliang22@m.fudan.edu.cn,yerong@bytedance.com},\\
\texttt{{\{xjhuang,zywei\}}@fudan.edu.cn},\\
\texttt{\{hanmeng12,lairuofei,zhangxinyu35\}@huawei.com}
}

\begin{document}
\maketitle
\begin{abstract}
How can we construct an automated debate judge to evaluate an extensive, vibrant, multi-turn debate? 
This task is challenging, as judging a debate involves grappling with lengthy texts, intricate argument relationships, and multi-dimensional assessments.
At the same time, current research mainly focuses on short dialogues, rarely touching upon the evaluation of an entire debate.
In this paper, by leveraging Large Language Models (LLMs), we propose Debatrix, which makes the analysis and assessment of multi-turn debates more aligned with majority preferences. 
Specifically, Debatrix features a vertical, iterative chronological analysis and a horizontal, multi-dimensional evaluation collaboration.
To align with real-world debate scenarios, we introduced the PanelBench benchmark, comparing our system's performance to actual debate outcomes.
The findings indicate a notable enhancement over directly using LLMs for debate evaluation.
Source code and benchmark data are available at \url{https://github.com/ljcleo/debatrix}.

\end{abstract}

\section{Introduction}
\label{sec:introduction}

Debating is the formal process of gaining consensus among groups with different opinions.
In many cases, such as in \emph{competitive} debates, only the policy proposed by the winning side will be accepted~\citep{Zhang2016Conversational}.
Debaters must apply various strategies to convince the audience to support their side in these debates.
While systems like Project Debater~\citep{Slonim2021autonomous} have enabled automatic \emph{debating} in competitive debates, \emph{judging} these debates still relies on human annotation.
Automating debate assessment is helpful to improve debate quality in political, commercial, or educational scenarios and can also accelerate the evolution of debate automatons.

Recently, large language models (LLM) such as ChatGPT and GPT-4~\citep{OpenAI2023GPT} have shown a solid ability to evaluate text quality \citep{Liu2023G,Chiang2023Can}.
In this task, LLMs can provide results more aligned with human preference than traditional metrics \citet{Zheng2023Judging}.
Additionally, LLM judges' verdicts are more straightforward to interpret than computational metrics or audience votes since they usually come with generated explanations.

\begin{figure}
    \centering
    \includegraphics[page=1,width=\linewidth]{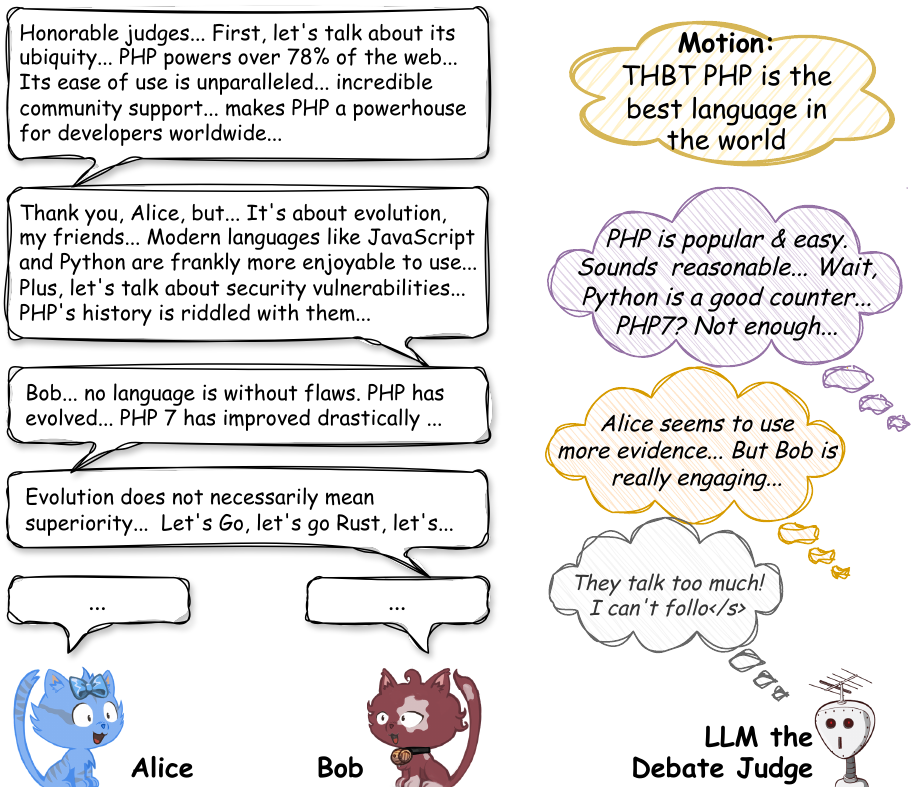}
    \caption{An LLM debate judge judging a debate between Alice and Bob. The LLM needs to understand the arguments and how they counter each other (purple bubble); the LLM also needs to evaluate the speeches in multiple dimensions (orange bubble). However, multi-round debates are often long, detracting attention or exceeding the context window (light gray bubble).}
    \label{fig:intro}
\end{figure}

However, judging debates with LLMs incorporates several issues to be considered, as illustrated in Figure~\ref{fig:intro}.
First, evaluating long, multi-turn debates remains challenging, as most current research focuses on short text such as open-question answers and user-request responses \citep{Zhong2022Towards,Wu2023Large}.
Second, debate judging requires a deep understanding of how arguments are organized and refuted across speeches;
not only does it require more professional knowledge, but it also demands a thorough, precise analysis of how arguments work in the debate.
Finally, speech quality is affected by various factors, such as argument strength, evidence reliability, and language style, requiring systematic evaluation across dimensions.

To this end, we propose Debatrix, a fine-grained framework to assist LLMs in handling these challenges by breaking down debate judging along both \emph{chronological} and \emph{dimensional} axes.
\begin{inparaenum}[1)]
    \item \textbf{Iterative Chronological Analysis}: We instruct the LLM to analyze the debate speech by speech, maintaining speech and analysis streams with a memory system and providing content analysis of previous speeches when analyzing new speeches. After reviewing all speeches, the LLM makes decisions based on all analyses. This iterative approach lets the LLM concentrate on one speech at a time and understand the context more effectively. It also produces feedback or decisions for each speech, each debater, and the final winner.
    \item \textbf{Multi-Dimensional Collaboration}: Debatrix also allows LLMs to focus on a specific judging dimension, such as argument, language, or clash, during the speech analyzing process. Each LLM agent can make comments on these specific aspects. For the overall judgment, all these individual analyses are combined into one summary, providing a systematic judgment across multiple dimensions.
\end{inparaenum}

Furthermore, we introduce PanelBench, a novel benchmark for evaluating automatic debate judging systems.
PanelBench consists of two collections of debates with judgments, namely DebateArt and BP-Competition.
DebateArt sources from online platforms that follow competitive debate formats; debates vary in speech count and length and have dimensional voting results.
BP-Competition includes debates transcribed from world-class competitive debate competitions;
these debates follow the British Parliamentary (BP) format involving four teams (two on each side), enriching PanelBench with long, complex, high-quality samples.
On PanelBench, Debatrix increased winner prediction accuracy compared to the baseline of directly prompting the LLM with raw speeches; per-dimension judging is also improved.
Furthermore, the results have also proved that iterative speech-by-speech analysis and splitting dimensions help generate a more accurate final verdict.

Our contributions are as follows:

\begin{enumerate}
    \item We propose Debatrix, a fine-grained automatic debate judging framework based on LLM, but performs analysis iteratively on multiple dimensions before summarizing and generating the final verdict.
    \item We propose PanelBench, a benchmark for automatic debate judging, including multi-dimensional and multi-debater settings, which differ from simple 1v1 debate assessment.
    \item We investigate how well LLMs can judge debates directly or using Debatrix, enabling chronological and/or dimensional analysis.
\end{enumerate}

\section{Related Work}
\label{sec:related}

Argumentation persuasion assessment is the foundation of automatic debate systems, as they must be persuasive enough to argue effectively.
Previous works have focused on the persuasiveness of arguments, including empirical studies~\citep{Thomas2017Adapting,Thomas2019Can}, machine learning models~\citep{Persing2015Modeling,Zhang2016Conversational,Gleize2019Are} and work covering both~\citep{AlKhatib2020Exploiting,Donadello2022Machine}.
Some argument-based chatbots also take persuasiveness as a motivating factor \citep{Rosenfeld2016Providing,Thomas2019Is}.
Few works have extended these methods to judging whole debates \citep{Potash2017Towards}, including competitive debates \citep{RuizDolz2023Automatic}.

While these works mainly involve empirical laws or delicate models, within the bigger context of text evaluation, large language models (LLMs) have become a new and powerful tool to tackle this task.
Multiple exploitation methods have been proposed, such as conditional probability~\citep{Fu2023GPTScore}, score prompting~\citep{Wang2023Is,Liu2023G,Zhu2023JudgeLM} and pairwise comparison~\citep{Wang2023PandaLM}.
Several works, such as \citet{Bai2023Benchmarking}, utilize multiple of them and introduce various methods to stabilize the results.

\begin{figure*}
    \centering
    \includegraphics[page=2,width=0.65\linewidth]{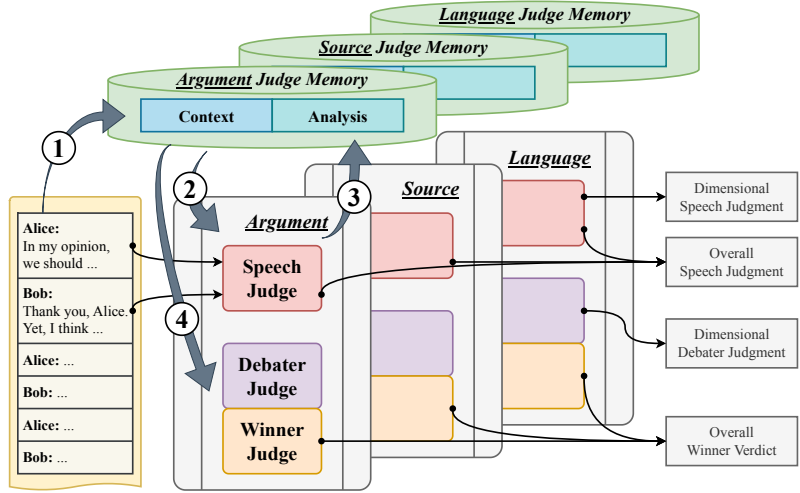}
    \caption{General structure of Debatrix. \ding{172}: add speech to context memory; \ding{173}: retrieve relevant pieces of context and analysis; \ding{174}: add analysis and reflections to analysis memory; \ding{175}: fetch analysis for final judgment. The framework can generate speech/debater judgments and the winner verdict based on analysis from single or multiple dimensions.}
    \label{fig:framework}
\end{figure*}

Regarding the challenges of judging debates with LLMs, some pioneering works have investigated some of them.
For instance, \citet{Chang2023Prompting}, \citet{Li2023PRD} and \citet{Chan2023ChatEval} focus on multi-dimensional assessment and propose different strategies to improve accuracy, such as dialetics, peer review and group chat.
Meanwhile, \citet{Wynter2023I} and \citet{Chen2023Exploring} have explored LLM's ability to handle argumentation tasks, measuring its capability of argumentation reasoning.
These works are partially in line with our work yet do not cover all \(3\) issues mentioned in Section~\ref{sec:introduction}.
Finally, \citet{Li2019ACUTE} proposed a dialog evaluation framework that works in a multi-turn manner, similar to our chronological design;
however, this framework is designed for human annotators instead of LLM judges.

\section{Debatrix}
\label{sec:framework}

In this section, we provide a detailed overview of Debatrix, our fine-grained debate judging framework.

\subsection{Structure and Components}

The overall structure of Debatrix is illustrated in Figure~\ref{fig:framework}.
Debatrix contains a collection of \textbf{iterative chronological analyses} like matrix columns, each capable of evaluating debates under a specific preference or dimension.
Each of them processes speeches one by one, using past analysis for new ones and generating judgments for every speech, including a score and a comment. 
When all speeches are analyzed, past analyses are summarized and converted into debater judgments (similar to speech judgments but for individual debaters) and the winner verdict (including the winner and a comment).
Multiple analyses can collaborate like a matrix, producing systematic judgments that cover multiple dimensions.

There are two components in Debatrix: memories and judges; they work together to analyze the debate thoroughly.
One iterative chronological analysis utilizes one memory and a set of judges.

\paragraph{Memory} Memories provide long-term storage during the debate judging process.
There are two types of memory: context memory records incoming speech context, and analysis memory stores intermediate analyses.
Every incoming speech is added to the context memory before being analyzed by the speech judge.
At any time, judges can fetch or query contents from both memories and add new analyses to the analysis memory.

\paragraph{Judge} Judges are the core components to analyze and judge the debate, including the speech judge, the debater judge, and the winner judge.
The speech judge analyzes the stream of speeches, utilizing memories to understand them;
the analysis is added to the memory and can be used to generate judgments.
The debater and winner judges work after all speeches are processed;
they use past analyses by the speech judges to generate debater judgments and winner verdicts, respectively.

\subsection{Iterative Chronological Analysis}

LLMs have shown a strong capability to evaluate single passages and short conversations.
To elaborate on this power for long multi-turn debates, we propose a dedicated design to analyze the debate iteratively in chronological order, which can already be seen in Figure~\ref{fig:framework}.
Figure~\ref{fig:design} illustrates a more detailed version; the complete algorithm is listed in Appendix~\ref{sec:algorithm}.

\begin{figure}
    \centering
    \includegraphics[page=3,width=0.6\linewidth]{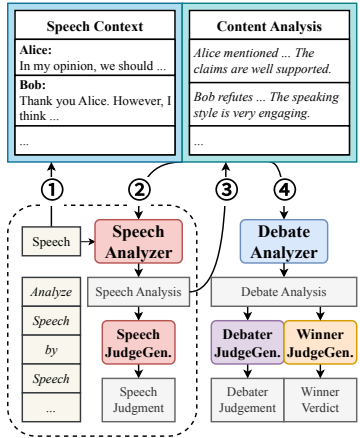}
    \caption{A more detailed version of an iterative chronological analysis. ``JudgeGen.''\ is a shorthand for ``judge/verdict generator'' that produces judgments and verdicts. Blocks and numbers match the ones in Figure~\ref{fig:framework}, except the debate analyzer not shown in Figure~\ref{fig:framework}.}
    \label{fig:design}
\end{figure}

The key point of the design is the iterative speech analysis process.
Speech analysis focuses on decomposing the content of the speech, such as how arguments interact with each other, what evidence is introduced to back arguments, and what language style the speech has shown.
The preference or dimension controls what is included in the analysis, forcing the speech judge to generate corresponding speech judgments.

Each speech analysis is added to the analysis memory as part of the content analysis, acting as a digested version of the current speech.
When analyzing the subsequent speech, we fetch both the new speech content and the content analysis of all previous speeches from the memory and input them to the LLM.
We can also include relevant speech contexts from previous speeches in the input.
This iterative approach reduces the complexity of the context and results in a more precise analysis.

At the end of the debate, the debater and the winner judges must take advantage of the entire list of content analyses to judge a specific debater or compare between debaters.
This is achieved by an extra debate analyzer, which converts all content analysis into a debater-directed debate analysis.
Finally, the debater judges and the winner judge generate debater judgments and the winner's verdict based on the debate analysis.

\subsection{Dimensional Collaboration}

One way to obtain the overall judgment of a debate --- a judgment covering all concerned dimensions --- is to configure a single iterative chronological analysis to consider multiple dimensions when analyzing speeches.
Yet, in Debatrix, we can also split and distribute dimensions to multiple analyses.
Each of them focuses on a specific dimension and can provide a more detailed insight regarding the dimension, which is useful when we demand dimensional comments on debates.

\begin{figure}
    \centering
    \includegraphics[page=4,width=0.55\linewidth]{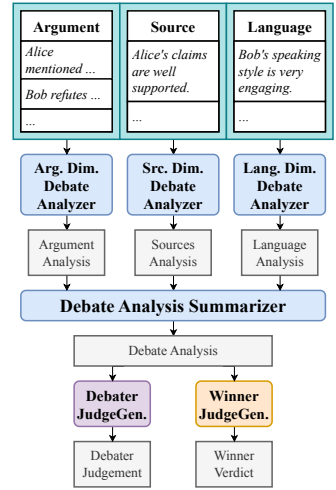}
    \caption{Combining multiple dimensional debate analyses into one systematic analysis at the end of the debate. Each dimension possesses a memory for dimensional content analysis, allowing a more nuanced understanding of the debate.}
    \label{fig:multidim}
\end{figure}

To synthesize an overall judgment from dimensional ones, Debatrix can combine debate analyses from multiple iterative chronological analyses into one systematic analysis.
Figure~\ref{fig:multidim} demonstrates the combination process for debater assessment and winner judging, summarizing debate analysis from various dimensions;
the same process can also be applied to speech analyses for speech judgments.
With improved dimensional analyses, Debatrix can generate more accurate overall judgments.

\section{PanelBench}
\label{sec:benchmark}

To assess LLMs and our proposed Debatrix framework with real competitive debates, we introduce PanelBench, a novel benchmark for automatic debate judging covering various styles, formations, and judging dimensions.
We include two debate collections in PanelBench: DebateArt for 1v1 debates with dimensional judgments and BP-Competition for high-quality debates with multiple debaters on each side.

\subsection{DebateArt}

DebateArt debates are collected from DebateArt\footnote{\url{https://debateart.org}}, an online debate platform that provides 1v1 debate arenas.
The formation setting makes the platform different from many other debate forums that do not restrict the speaking order, as debates on this platform are closer to formal competitive ones.

In DebateArt, users vote to decide debate winners.
Besides the common winner selection system, the platform also provides a categorical point assignment system, where voters must consider and vote on four metrics for comparative performance insights:
Argument, source, legibility, and conduct.\footnote{\url{https://info.debateart.com/terms-of-service/voting-policy\#casting-votes}}
This voting system provides judgments under separate dimensions: pro-winning, con-winning, or a tie.
Moreover, voters must provide detailed explanations of their decisions, and their votes are supervised by experienced moderators, enhancing their quality.
The weighted average of dimensional votes decides the debate's overall winner.
Appendix~\ref{subsec:debateart} details how DebateArt debates are run.

\begin{table}
    \centering
    \begin{tabular}{l@{}rrr}
\toprule
     & \# Speech & Speech Tok. &  Debate Tok. \\ \midrule
Min  & \(4.0\)   & \(53.0\)    & \(468.0\)    \\
Mean & \(6.7\)   & \(650.5\)   & \(4,342.6\)  \\
Max  & \(10.0\)  & \(2,368.0\) & \(12,337.0\) \\ \bottomrule
\end{tabular}
    \caption{DebateArt debates content statistics, including the number of speeches in a debate, tokens in a speech, and tokens in a debate.}
    \label{tab:dart_content_stat}
\end{table}

\begin{table}
    \centering
    \begin{tabular}{lrrrr}
\toprule
Dimension & Pro    & Tie    & Con    & D2O RMSE  \\ \midrule
Overall   & \(37\) & \(7\)  & \(56\) & \(-\)     \\ \midrule
Argument  & \(33\) & \(11\) & \(56\) & \(23.85\) \\
Source    & \(14\) & \(67\) & \(19\) & \(41.02\) \\
Language  & \(9\)  & \(66\) & \(25\) & \(47.20\) \\ \bottomrule
\end{tabular}
    \caption{DebateArt debates winner distribution, including overall and dimensional ones. The rightmost column is the root mean square error (RMSE, 100x) when matching dimensional winners to overall ones by assigning pro, tie, and con to \(0\), \(0.5\), and \(1\), respectively.}
    \label{tab:dart_winner_stat}
\end{table}

PanelBench includes \(100\) valid debates with valid votes from DebateArt;
we include our data collection process and filtering criteria in Appendix~\ref{subsec:debateart-col}.
Table~\ref{tab:dart_content_stat} lists the statistics of these debates.
To align with oral debates that are not formatted, we merged two dimensions --- legibility and conduct --- into a single language dimension, representing the language style.
We averaged their votes in these two dimensions for each vote and converted them into a single vote (pro/tie/con).
Table~\ref{tab:dart_winner_stat} lists the winner distribution of all debates:
While voters tend to give ties to source and language, most debates have a specific winning side, largely because of the argument dimension.

\subsection{BP-Competition}

To extend PanelBench with high-quality \textit{formal} debates, we furthermore collected debates from recent world-class competitive debate competitions,
including the World Universities Debating Championship~(WUDC), the European Universities Debating Championship~(EUDC), and the North American Debating Championship~(NAUDC).
All these competitions follow the British Parliamentary~(BP) format~\citep{UniWorldDebatingCouncil2023World}; therefore, we name them \emph{BP-Competition} debates.
In this format, four teams (OG, OO, CG, and CO) are divided into two sides of a motion but compete against all three other teams in the debate;
hence, instead of predicting the winning side, PanelBench requires judging which of the four teams is the best.
Please refer to Appendix~\ref{subsec:bpdebate} for more details about the BP format.

\begin{table}
    \centering
    \begin{tabular}{l@{}rr}
\toprule
     & Speech Tok. &  Debate Tok. \\ \midrule
Min  & \(1,478.0\) & \(13,571.0\) \\
Mean & \(1,892.5\) & \(15,139.9\) \\
Max  & \(2,411.0\) & \(17,089.0\) \\ \bottomrule
\end{tabular}
    \caption{BP-Competition debates content statistics, including the number of tokens in a speech and a debate.}
    \label{tab:bp_content_stat}
\end{table}

\begin{table}
    \centering
    \begin{tabular}{lrrrr}
\toprule
Debater & OG    & OO     & CG    & CO    \\ \midrule
\# Wins & \(8\) & \(16\) & \(8\) & \(6\) \\ \bottomrule
\end{tabular}
    \caption{BP-Competition debates winner distribution. Debater names are fixed to OG, OO, CG, and CO; OG and CG form the pro side, and OO and CO form the con side. Some debates have two winners, so the sum exceeds \(22\).}
    \label{tab:bp_winner_stat}
\end{table}

We transcribed debate videos from knockout rounds of famous competitive debate competitions to obtain high-quality BP debates;
Please refer to Appendix~\ref{subsec:bpdebate-col} for data collection details.
After filtering incomplete or damaged transcriptions, we obtained \(22\) debates with full transcriptions and final verdicts.
Table~\ref{tab:bp_content_stat} lists statistics of these debates; they are significantly longer than DebateArt ones.

It is worth noting that we were only able to collect the winning teams of BP-Competition debates, which is not necessarily unique:
In the semifinals and quarter-finals, two of them (they can even be mutual opponents) can win and proceed.
Among all BP-Competition debates, \(6\) have only one winner, and \(16\) have two;
Table~\ref{tab:bp_winner_stat} lists the distribution of all winners.
To unify them, PanelBench treats predicting any winning teams as correct.

\section{Experiments}
\label{sec:experiment}

We conducted experiments on PanelBench to evaluate the debate-judging performance of LLMs.
We also compare our Debatrix framework with judging directly with LLMs.

\subsection{Model \& Framework Configuration}

We utilize the latest GPT family as our target LLMs, including ChatGPT and GPT-4\footnote{For ChatGPT, we use \texttt{gpt-3.5-turbo-0125}; for GPT-4, we use \texttt{gpt-4-0125-preview}.};
our experiments mainly focus on ChatGPT to test Debatrix under a limited context window (\(16,385\) tokens).
We set the temperature to 0 and repeated all experiments \(3\) times, measuring the average performance.

We provide judge preferences in the system prompt to control the dimension of the judgments.
To shorten the input, we do not include relevant context from previous speeches, only the content analysis.
We ask the judges to output comments only and then call the LLM again to generate their corresponding scores (integers from \(1\) to \(10\)) or winner verdict to diminish mismatches.
All prompt templates are listed in Appendix~\ref{sec:prompt}.

As mentioned in Section~\ref{sec:benchmark}, we introduce \(3\) judging dimensions for DebateArt debates: argument, source, and language.
In our experiments for these debates, we stick to these dimensions and their weights (argument: 3; source: 2; language: 2) to measure dimensional judging performance and utilize them to generate systematic verdicts.
BP-Competition debates do not have dimensional judgments, yet we continue splitting dimensions for these debates.
We adopt the dimensions defined for DebateArt and add a clash dimension, which originally belongs to the argument dimension, so that all dimensions are balanced (and thus, all dimensions have equal weight).

\subsection{Baseline Models}

We compare Debatrix with these baseline models:

\begin{itemize}
    \item \textbf{ChatGPT} and \textbf{GPT-4} reads the entire debate in one input and generates an overall verdict, with an all-in-one judge preference covering all dimensions, backbone LLMs being ChatGPT and GPT-4 respectively;
    \item \textbf{Chronological} introduces iterative chronological analysis but does not split dimensions, instead using the all-in-one judge preference.
    \item \textbf{Dimensional} introduces dimensional collaboration, assessing debates under every dimension and summarizing dimensional judgments, but inputs the debate as a whole;
    \item \textbf{NonIterative} is almost identical to Debatrix, but the analysis is not iterative: we fetch the \emph{raw content} instead of the content analysis of all previous speeches when analyzing new speeches, so the speech analysis is solely based on the original text.
\end{itemize}

We use \texttt{ChatGPT}, \texttt{GPT-4}, \texttt{Debatrix}, etc.\ to refer to the models below, distinguishing them from backbone LLMs and frameworks.

\subsection{Evaluation Configuration}

We generate winner verdict predictions for both debate collections according to the model output and compare them with true verdicts.
We experiment with two methods of winner prediction for DebateArt debates:
\textbf{score comparison} (SC) compares the score of each debater generated by debater judges and chooses the debater with a higher score (or announces a tie when the scores are equal);
\textbf{direct prediction} (DP) predicts the winner directly according to the verdicts by the winner judges.
Specifically, for the source and language dimensions, we treat score differences within \(\pm3\) as ties, as human users tend to neglect small differences in them (as can be seen in Table~\ref{tab:dart_winner_stat}).
We only utilize direct prediction for BP-Competition debates since they do not allow ties and score comparison often fails to produce a single winner.

\begin{table*}
    \centering
    \begin{tabular}{lcccc}
\toprule
\multirow{2}{*}{Model} & \multicolumn{2}{c}{DebateArt}                   & \multicolumn{2}{c}{BP-Competition}                   \\ \cmidrule(l){2-5} 
                       & SC RMSE \(\downarrow\) & DP RMSE \(\downarrow\) & Completion Rate \(\uparrow\) & Accuracy \(\uparrow\) \\ \midrule
\texttt{ChatGPT}       & \(49.99\)              & \(51.16\)              & \(13.64\)                    & \(0.00\)              \\
\texttt{GPT-4}         & \(44.84\)              & \(46.55\)              & \(\mathbf{100.00}\)          & \(34.85\)             \\ \midrule
\texttt{Chronological} & \(48.61\)              & \(48.71\)              & \(\mathbf{100.00}\)          & \(30.30\)             \\
\texttt{Dimensional}   & \(44.91\)              & \(45.01\)              & \(36.36\)                    & \(12.12\)             \\
\texttt{NonIterative}  & \(44.18\)              & \(44.03\)              & \(66.67\)                    & \(36.36\)             \\ \midrule
\texttt{Debatrix}      & \(\mathbf{42.21}\)     & \(\mathbf{41.75}\)     & \(\mathbf{100.00}\)          & \(\mathbf{51.52}\)    \\ \bottomrule
\end{tabular}
    \caption{Debate judging performance on PanelBench. SC is score comparison; DP is direct prediction. All metrics are enlarged to 100x. Lower RMSE and higher completion rate/accuracy are better; bold font indicates best results.}
    \label{tab:exp_main}
\end{table*}

When evaluating model performance on DebateArt debates, we measure the root mean square error (RMSE) between the model prediction and the true verdict due to the existence of ties.
More specifically, we assign the values of pro, tie, and con verdicts to \(0\), \(0.5\), and \(1\), respectively;
then, we match each true verdict to its corresponding model prediction and calculate the RMSE.
Meanwhile, we directly measure the completion rate and prediction accuracy for BP-Competition debates since some models fail to generate complete verdicts because of a limited context window.

\subsection{Main Results}

Table~\ref{tab:exp_main} lists the main experiment results across models and debate collections.
Appendix~\ref{subsec:significance} provides significance tests of the results.
\texttt{Debatrix} consistently outperforms all baselines on both debate collections, including \texttt{GPT-4}.

Iterative chronological analysis is crucial for ChatGPT, with a limited context window, to handle very long debates:
only \texttt{Chronological} and \texttt{Debatrix} judged all \(22\) BP-Competition debates in all trials successfully, while \texttt{ChatGPT} only completed very few judgments and none of them are correct;
\texttt{NonIterative} failed to judge some BP-Competition debates because raw speeches are much longer than content analyses.

Meanwhile, dimensional collaboration is also beneficial in shorter debates, as seen in the DebateArt collection.
When debates are short enough to fit in the context window, \texttt{Dimensional} significantly outperforms \texttt{ChatGPT} and is on par with \texttt{GPT-4}.
On BP-Competition debates, the completion rate also increases since judging a single dimension costs fewer tokens.

Nevertheless, combining both of them gives the best performance.
\texttt{NonIterative} is already aligned with \texttt{GPT-4}, especially superseding it on the BP-Competition collection while not even completing all trials.
However, using previous content analysis iteratively, \texttt{Debatrix} overcomes the context window issue and shows significant improvements compared to \texttt{NonIterative}.

\section{Analysis}
\label{sec:analysis}

\subsection{Multi-Turn \& Long Debates}

\begin{figure*}
    \centering
    \includegraphics[page=1,width=\linewidth]{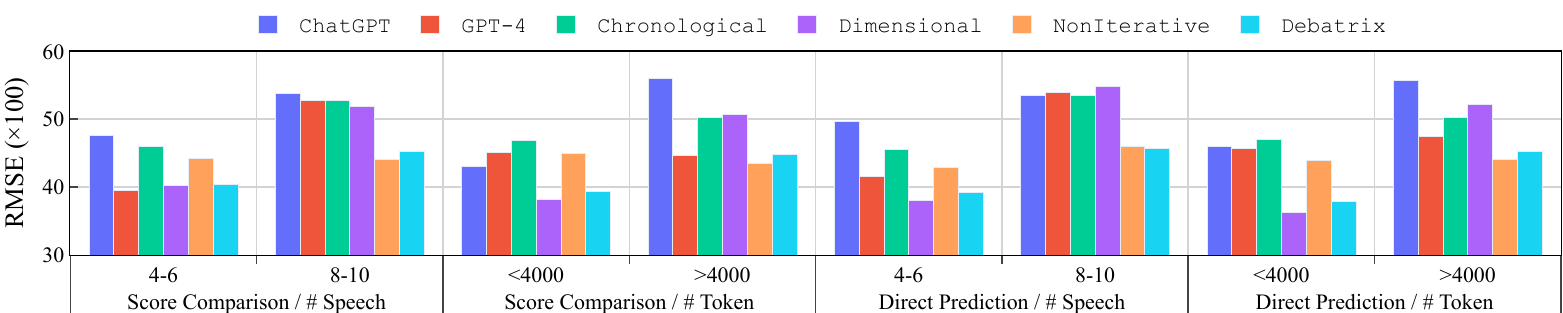}
    \caption{Winner prediction RMSE on DebateArt debates with different numbers of speeches (left) or tokens (right). Note that \texttt{Debatrix} (light blue bar) always reaches a relatively low RMSE regardless of the number of speeches or tokens; no other model can achieve this.}
    \label{fig:exp_long}
\end{figure*}

As demonstrated in Table~\ref{tab:dart_content_stat}, DebateArt debates vary in speech count and length:
among all \(100\) debates, \(34\) have no fewer than \(8\) speeches, and \(51\) contain at least \(4,000\) tokens.
Figure~\ref{fig:exp_long} compares the performance of different models on debates with different numbers of speeches or tokens.

On one hand, some baseline models show partial advantage but fail to cover all scenarios:
\texttt{GPT-4} handles long debates well but degrades when judging debates with many turns;
\texttt{Dimensional} achieves low RMSE on debates with few turns or tokens and \texttt{NonIterative} on ones with many turns or tokens, but both become rather worse when switching to the other half of the debates.

On the other hand, \texttt{Debatrix} maintains a relatively low RMSE compared to other models, regardless of the number of speeches or tokens.
This shows that Debatrix effectively assists the LLM in evaluating long, multi-turn debates while extending its advantage to short debates.

\subsection{Performance on Argument Dimension}

\begin{table}
    \centering
    \begin{tabular}{lccc}
\toprule
Model                        & SC                 & DP                 \\ \midrule
\texttt{Dimensional}         & \(52.06\)          & \(52.23\)          \\
\texttt{Dimensional} (GPT-4) & \(51.13\)          & \(51.06\)          \\
\texttt{NonIterative}        & \(50.87\)          & \(50.37\)          \\ \midrule
\texttt{Debatrix}            & \(\mathbf{47.50}\) & \(\mathbf{47.67}\) \\ \bottomrule
\end{tabular}
    \caption{Winner prediction RMSE on DebateArt debates under the argument dimension. SC is score comparison; DP is direct prediction. All metrics are enlarged 100x; bold font indicates the best results.}
    \label{tab:exp_dart_arg}
\end{table}

Among all the dimensions we introduced, the argument dimension is the major one affecting debaters' persuasiveness, as shown in Table~\ref{tab:dart_winner_stat}.
Table~\ref{tab:exp_dart_arg} lists model performance on DebateArt debates under this dimension.
We only include models with dimensional collaboration;
we also modify \texttt{Dimensional} to use GPT-4 to examine whether upgrading the LLM is beneficial.

We can see in Table~\ref{tab:exp_dart_arg} that using a more powerful LLM such as GPT-4 does not improve \texttt{Dimensional} much on this dimension;
instead, chronological analysis brings a significant improvement.
Iterative analysis when analyzing speeches is also beneficial, as illustrated by the difference between \texttt{Debatrix} and \texttt{NonIterative}.
All these results prove that Debatrix can help the LLM understand arguments better without resorting to larger models.

\subsection{Dimensional Collaboration for Systematic Judgment}

Table~\ref{tab:exp_main} also clearly demonstrates the advantage of dimensional collaboration.
Models that split dimensions (\texttt{Dimensional} and \texttt{Debatrix}) consistently outperform their duets without splitting dimensions (\texttt{ChatGPT} and \texttt{Chronological}).
This shows that splitting dimensions improves the analysis quality for each dimension, leading to a more precise systematic judgment at the end.
Specifically, \texttt{Dimensional} significantly outperforms \texttt{Chronological} on DebateArt debates, indicating that the quality of intermediate content analysis is crucial for the final judgment.

\subsection{Position Bias on GPT-4}

\begin{table}
    \centering
    \begin{tabular}{lr@{}r@{\hspace\tabcolsep}r@{\hspace\tabcolsep}r@{\hspace\tabcolsep}r}
\toprule
\multirow{2}{*}{Model} & \multicolumn{2}{c}{Opening} & \multicolumn{2}{c}{Closing} & \multirow{2}{*}{N/A} \\ \cmidrule(lr){2-5}
                       & OG     & OO                 & CG       & CO               &                      \\ \midrule
\texttt{GPT-4}         & \(0\)  & \(0\)              & \(10\)   & \(56\)           & \(0\)                \\
\texttt{NonIterative}  & \(17\) & \(12\)             & \(9\)    & \(6\)            & \(22\)               \\
\texttt{Debatrix}      & \(13\) & \(21\)             & \(15\)   & \(17\)           & \(0\)                \\ \midrule
\emph{Gold}            & \(12\) & \(30\)             & \(13.5\) & \(10.5\)         & \(0\)                \\ \bottomrule
\end{tabular}
    \caption{Distribution of winner predictions on BP-Competition debates. \emph{Gold} is the expectation number each debater is chosen if we select one of the true winners randomly for each trial. Note that the opening half (OG and OO) speaks before the closing (CG and CO).}
    \label{tab:exp_bp_dist}
\end{table}

From Table~\ref{tab:exp_main}, we can find that \texttt{GPT-4}, while having a context window large enough to handle all BP-Competition debates, does not achieve a high accuracy.
To investigate what prevents \texttt{GPT-4} from better performance, Table~\ref{tab:exp_bp_dist} summarizes the winner prediction distribution of various models.
Surprisingly, \texttt{GPT-4} always predicts the closing half (CG and CO), which speaks after the opening half (OG and OO); in most cases, it selects CO, who speaks the last.
Meanwhile, ChatGPT-based \texttt{Debatrix} gives relatively balanced predictions roughly matching the expectation.
The case study in Appendix~\ref{subsec:case} also reveals this phenomenon.

We conjecture that position bias \citep{Ko2020Look,Wang2023PandaLM} could be a significant factor that causes GPT-4 to fail in judging BP debates.
When the arguments of all debaters are similarly strong, LLM may prefer the last speaker who can refute other debaters without being refuted by others, thus seemingly more convincing.

\section{Conclusion}
\label{sec:conclusion}

In this paper, we propose a fine-grained debate judging framework based on LLM, Debatrix.
We decompose the debate judging task into an iterative chronological analysis to tackle multi-turn, long debates and elaborate multiple dimensions to generate systematic judgments.
We introduce a novel debate judging benchmark, PanelBench, to assess our framework and other automatic debate judging approaches.
The benchmark covers multi-dimensional and multi-debater scenarios.
Under both settings, Debatrix significantly improves ChatGPT, aiding it in judging long debates that exceed the context window and outperforming GPT-4.

\section*{Limitations}
Despite the above results, this paper has a few limitations for which we appreciate future studies.
First, while the iterative chronological analysis in Debatrix already shows competence, it can be further studied and polished.
Second, the position bias issue on GPT-4 for BP-Competition debates needs further investigation, and more powerful tools are needed to rectify this phenomenon.

\section*{Ethical Impact}
As a debate judging framework based on LLM, Debatrix may be exposed to ethical risks if the backbone LLM is biased beforehand, especially when judging debates concerning political, social, or cultural motions.
In our experiments, however, using LLMs with strong ethical constraints like ChatGPT and GPT-4 has resulted in no significant ethical bias within the generated verdicts.
Furthermore, as demonstrated in Appendix~\ref{subsec:case}, Debatrix guides LLMs to focus on the content instead of basing their judgments on a priori knowledge or belief, alleviating potential ethical impact.

\section*{Acknowledgements}
This work is supported by National Natural Science Foundation of China (No. 62176058) and National Key R \& D Program of China (2023YFF1204800). The project's computational resources are supported by CFFF platform of Fudan University.

\bibliography{main}
\bibliographystyle{acl_natbib}

\appendix

\section{Debatrix Algorithm}
\label{sec:algorithm}

\begin{algorithm*}
    \small
    \caption{Debatrix Iterative Chronological Analysis}\label{alg:framework}
    \SetKwData{SCM}{sc\_mem}
    \SetKwData{CAM}{ca\_mem}
    \SetKwData{CI}{ci}
    \SetKwData{RSC}{rel\_sc}
    \SetKwData{CA}{ca}
    \SetKwData{ACA}{all\_ca}
    \SetKwData{DA}{da}
    \SetKwFunction{QSC}{QuerySpeechContext}
    \SetKwFunction{AC}{AnalyzeContent}
    \SetKwFunction{JS}{JudgeSpeech}
    \SetKwFunction{FCA}{FetchContentAnalysis}
    \SetKwFunction{AD}{AnalyzeDebate}
    \SetKwFunction{JD}{JudgeDebater}
    \SetKwFunction{JW}{DecideWinner}
    \SetKwInOut{Input}{input}\SetKwInOut{Output}{output}

    \Input{judge preference \(P\), debate motion \(M\), a list of debaters \(\{D_1, \dots, D_m\}\), an info slide \(I\), a list of speaker-speech tuples \(\{(d_1, s_1), \dots, (d_n, s_n)\}\)}
    \Output{a list of speech comments \(\{C_{s_1}, \dots, C_{s_n}\}\), a list of debater comments \(\{C_{D_1}, \dots, C_{D_m}\}\), a winner verdict \(V\)}
    \BlankLine
    \makebox[4em][l]{\SCM} \(\leftarrow \emptyset\)\tcp*{speech context memory}
    \makebox[4em][l]{\CAM} \(\leftarrow \emptyset\)\tcp*{content analysis memory}
    \makebox[4em][l]{\CI}  \(\leftarrow (P, M, \{D_1, \dots, D_m\}, I)\)\tcp*{common inputs}
    \BlankLine
    \For{\(i \leftarrow 1\) \KwTo \(n\)}{
        \makebox[4em][l]{\RSC}        \(\leftarrow\) \QSC{\SCM, \(s_i\)}\tcp*{query relevant contents}
        \makebox[4em][l]{\SCM}        \(\leftarrow\) \SCM \(\cup \{(d_i, s_i)\}\)\;
        \makebox[4em][l]{\CA}         \(\leftarrow\) \AC{\CI, \(d_i\), \(s_i\), \CAM, \RSC}\tcp*{analyze speech}
        \makebox[4em][l]{\CAM}        \(\leftarrow\) \CAM \(\cup\) \CA\;
        \makebox[4em][l]{\(C_{s_i}\)} \(\leftarrow\) \JS{\CI, \(d_i\), \CA}\tcp*{judge speech}
    }
    \BlankLine
    \makebox[3em][l]{\ACA}  \(\leftarrow\) \FCA{\CAM}\tcp*{fetch all content analyses}
    \makebox[3em][l]{\DA}   \(\leftarrow\) \AD{\CI, \ACA}\tcp*{analyze debate}
    \BlankLine
    \For{\(i \leftarrow 1\) \KwTo \(m\)}{
        \makebox[2em][l]{\(C_{D_i}\)} \(\leftarrow\) \JD{\CI, \(D_i\), \DA}\tcp*{judge debater}
    }
    \BlankLine
    \makebox[3em][l]{\(V\)} \(\leftarrow\) \JW{\CI, \DA}\tcp*{decide winner of debate}
    \Return{\(\{C_{s_1}, \dots, C_{s_n}\}, \{C_{D_1}, \dots, C_{D_m}\}, V\)}
\end{algorithm*}

\subsection{Iterative Chronological Analysis Algorithm}

Algorithm~\ref{alg:framework} demonstrates the complete algorithm of iterative chronological analysis in Debatrix.
Here, we do not include dimensional collaboration.
Note that our experiments do not query and include relevant contexts from previous speeches.

\subsection{Computational Cost Analysis}

Here, we provide an approximate analysis of Debatrix's computational cost based on the number of input tokens.
We do not consider the number of generated tokens because, first, they are much fewer than the input tokens (especially for long debates) and have much less impact; second, extra outputs from Debatrix can be further utilized, for example, when generating speech verdicts.

Assume we have a debate with \(N\) speeches, each speech having \(x\) tokens.
Furthermore, assume each speech analysis has \(y\) tokens and debate analysis \(z\) tokens.
We follow our experiment settings in Section~\ref{sec:experiment} for all estimations.

First, we consider the single-dimension case.
\texttt{ChatGPT} or \texttt{GPT-4} consumes \(Nx\) input tokens and generates \(z\) output tokens.
\texttt{Debatrix} consumes \(y\cdot N(N+1)/2\) more input tokens and generates \(Ny\) more output tokens because we input the speech analysis for all later speeches.
For long debates, for example BP-Competition where \(N=8\) and \(y\approx x/6\), the number of input tokens is only \(1.75\) times more than \texttt{ChatGPT}/\texttt{GPT-4}.
Even for debates with shorter speeches, like DebateArt where \(y\approx x/2\), the ratio is less than \(4\).

Now we turn to dimensional collaboration with \(k\) dimensions.
In this case, the total cost is multiplied by \(k\), with an extra \(kz \to z\) step to generate the summarized judgment.
In our experiments, \(k=3\) (DebateArt) or \(k=4\) (BP-Competition) and \(z\approx y\).
This results in a final ratio lower than \(20\) --- the price ratio between GPT-4 and ChatGPT\footnote{\url{https://openai.com/pricing}}.

In Section~\ref{sec:experiment}, we show that ChatGPT-based \texttt{Debatrix} consistently outperforms \texttt{GPT-4}; the above estimation demonstrates that our computational cost is also lower.

\section{Benchmark Data Details}
\label{sec:source}

\subsection{DebateArt Debate Procedure\protect\footnote{This section mainly refers to \url{https://info.debateart.com/help/debates}.}}\label{subsec:debateart}

Users instigate debates in DebateArt. The instigator needs to provide a debate topic and set debate configurations such as character limit, time limit (12 hours to 2 weeks), and the number of rounds (up to 5); they can also include a description with pertinent details like definitions, expanded resolution, special rules, and scope limitations.

The instigator may elect to be pro or con, leaving the other position to the contender; the contender can be any community user willing to accept the challenge or another user requested directly by the instigator. No matter which case, once the contender enters, the debate starts, and both sides publish their arguments. If any side fails to propose an argument within the time limit, they will automatically forfeit the round; in our work, debates with forfeited turns are treated as incomplete.

When all arguments have been published, the community or the appointed judges select the debate's winner by voting. Voters need to follow the specified voting system and give fair verdicts. The debate is finished when the winner has been selected according to the votes.

\subsection{DebateArt Debate Collection}\label{subsec:debateart-col}

To collect debates with valid content and votes, we first crawled the list of finished debates on DebateArt and then filtered out debates that had no valid votes or were interrupted (not using all preset rounds). Next, we crawled the debate details of the remaining debates, including topic (motion), debaters, description (info slide), arguments (speeches), and votes. The raw arguments are in HTML format; we use \texttt{markdownify}\footnote{\url{https://github.com/matthewwithanm/python-markdownify}} to convert them into Markdown documents.

We further filtered out debates that do not fit our benchmark, including debates that do not use the categorical point assignment system, are not formal, and contain very short speeches, which may indicate a forfeit. Although many are close to being professionals, we also excluded some very long debates to ensure that inputs do not exceed the LLM's context window during the experiment.

\subsection{BP Debates}\label{subsec:bpdebate}

The British Parliamentary (BP) format is a widely accepted competitive debate style, followed by famous competitions like WUDC, EUDC, and NAUDC \citep{UniWorldDebatingCouncil2023World}. Each BP debate contains four teams, with a total of eight debaters. There are two teams on each side of the debate: on one side are Opening Government (OG) and Closing Government (CG); on the other side are Opening Opposition (OO) and Closing Opposition (CO). They follow the order specified below to give speeches:

\begin{itemize}
    \item OG Speaker 1 (``Prime Minister'');
    \item OO Speaker 1 (``Leader of Opposition'');
    \item OG Speaker 2 (``Deputy Prime Minister'');
    \item OO Speaker 2 (``Deputy Leader of Opposition'');
    \item CG Speaker 1 (``Government Member'');
    \item CO Speaker 1 (``Opposition Member'');
    \item CG Speaker 2 (``Government Whip'');
    \item CO Speaker 2 (``Opposition Whip'').
\end{itemize}

Each speech lasts for \(7\) minutes, with limited tolerance for timeouts. In general, OG should define the motion, propose arguments, and refute arguments from OO; OO should rebut OG's case and propose constructive arguments for their side; CG and CO should provide further supplementary analysis in favor of their side, respectively.

Points of Information (POI) is a special feature of BP debates. A POI is a formalized interjection from any debater on the opposite side to the current speaker. The current speaker has the right to decide whether the POI is accepted or rejected; once accepted, the debater offering the POI can make an argument or ask a question within 15 seconds, and the current speaker should respond properly before continuing their speech. We mark POI conversations as quoting blocks and prompt LLMs to pay attention to them, as engaging in POIs may contribute to the debaters' overall performance.

\subsection{Competition Debate Collection}\label{subsec:bpdebate-col}

Many debate competitions, including WUDC, EUDC, and NAUDC, only provide essential information about debates, such as motions, info slides, teams, and winners. They do not have official transcriptions; unofficial ones are often incomplete. Fortunately, in recent years, many of these competitions have provided official video recordings of debates not long before the finals.

We selected debates starting from the quarter-finals from WUDC (2020-2023), EUDC (2019-2022), and NAUDC (2021 and 2023) and downloaded their video recordings. Next, we extracted audio files from the recordings and used Whisper \citep{Radford2023Robust} to recognize the speeches. We manually checked and formatted the output results into valid transcriptions.

Not every debate was available for transcription due to missing, damaged, or incomplete recordings; we only kept debates whose transcription was complete. Finally, we merged their transcriptions with debate information to produce the final data.

\section{Prompt Template}
\label{sec:prompt}

Tables~\ref{tab:pr_dart_update},~\ref{tab:pr_dart_judge}~and~\ref{tab:pr_dart_summarize} demonstrates the prompt templates in our experiments.
For BP-Competition debates, we insert an extra section describing BP debate characteristics in the system message right after the debate information section, as shown in Table~\ref{tab:pr_bp_desc}.
Dimensional preferences of each stage are available in the PanelBench release.

\begin{table*}
    \centering\small
    \begin{tabular}{l}
\toprule
\textbf{System:} \\ \midrule
\begin{minipage}{0.95\linewidth}
As an AI with expertise in competitive debating, you're serving as a judge on a panel.

The debate motion is: \emph{<debate motion>}

Pro side debaters are: \emph{<pro debaters>}

Con side debaters are: \emph{<con debaters>}

The debate consists of \emph{<speech count>} speeches.

Now, \emph{<new speech speaker>} gives Speech \emph{<new speech id>}. You are given the debate info slide. Also, you have analyzed all previous speeches made in the debate.

\emph{<content analysis dimensional preference>}

Please think critically before responding.
\end{minipage} \\ \midrule
\textbf{User:} \\ \midrule
\begin{minipage}{0.95\linewidth}
\# Info Slide

\emph{<info slide>}

\# Your Analysis of Speech 1 by \emph{<speech 1 speaker>}

\emph{<speech 1>}

(more speech analyses \ldots)

\# New Speech by \emph{<new speech speaker>}

\emph{<new speech>}
\end{minipage} \\ \bottomrule
\end{tabular}
    \caption{Speech analyzer prompt template for \texttt{Debatrix} on DebateArt debates. Prompt sections about previous speeches are included only when there \emph{are} previous speeches; for \texttt{NonIterative}, analyses are replaced with raw speeches.}
    \label{tab:pr_dart_update}
\end{table*}

\begin{table*}
    \centering\small
    \begin{tabular}{l}
\toprule
\textbf{System:} \\ \midrule
\begin{minipage}{0.95\linewidth}
As an AI with expertise in competitive debating, you're serving as a judge on a panel.

The debate motion is: \emph{<debate motion>}

Pro side debaters are: \emph{<pro debaters>}

Con side debaters are: \emph{<con debaters>}

The debate consists of \emph{<speech count>} speeches.

Now the debate ends, and you have analyzed all speeches made in the debate.

\emph{<debate analysis dimensional preference>}

Judge the \emph{<debater count>} debaters (\emph{<debaters>}) individually. Ties are possible, so if more than one debater gives the best performance, announce a tie, otherwise award one and only one individual debater that performs the best.

Please think critically before responding.
\end{minipage} \\ \midrule
\textbf{User:} \\ \midrule
\begin{minipage}{0.95\linewidth}

\# Your Analysis of Speech 1 by \emph{<speech 1 speaker>}

\emph{<speech 1>}

(more speech analyses \ldots)
\end{minipage} \\ \bottomrule
\end{tabular}
    \caption{Debate analyzer prompt template for \texttt{Debatrix} on DebateArt debates. For \texttt{ChatGPT}, \texttt{GPT-4} and \texttt{Dimensional}, analyses are replaced by raw speeches, and the info slide is included in the user message like in Table~\ref{tab:pr_dart_update}. If the current dimension does not allow ties, ''Ties are possible \ldots, otherwise award \ldots'' is replaced by ``Ties are not allowed, so you should award \ldots''}
    \label{tab:pr_dart_judge}
\end{table*}

\begin{table*}
    \centering\small
    \begin{tabular}{l}
\toprule
\textbf{System:} \\ \midrule
\begin{minipage}{0.95\linewidth}
As an AI with expertise in competitive debating, you're serving as the main judge on a panel.

The debate motion is: \emph{<debate motion>}

Pro side debaters are: \emph{<pro debaters>}

Con side debaters are: \emph{<con debaters>}

The debate consists of \emph{<speech count>} speeches.

Now the debate ends. Your team of specialized judges, each assigned to a dimension, have given their judgments according to their assigned dimension respectively. You are given the dimensions and corresponding judgments from your team.

Please summarize the dimensional judgments according to their weights into a general judgment. Rank the \emph{<debater count>} debaters (\emph{<debaters>}) individually; award one and only one individual debater that performs the best.

Please think critically before responding.
\end{minipage} \\ \midrule
\textbf{User:} \\ \midrule
\begin{minipage}{0.95\linewidth}
\# \emph{<dimension 1 name>}

Weight: \emph{<dimension 1 weight>}

\emph{<dimension 1 dimensional judgment>}

(more dimensions \ldots)
\end{minipage} \\ \bottomrule
\end{tabular}
    \caption{Debate analysis summarizer prompt template for \texttt{Debatrix} on DebateArt debates.}
    \label{tab:pr_dart_summarize}
\end{table*}

\begin{table*}
    \centering\small
    \begin{tabular}{l}
\toprule
\textbf{System:} \\ \midrule
\begin{minipage}{0.95\linewidth}
\ldots

The debate consists of \emph{<speech count>} speeches.

This is a British Parliamentary style debate: the debaters' names are abbreviations of Opening/Closing Government/Opposition respectively; OG and OO first give 4 speeches in turn, then CG and CO give another 4 speeches in turn.

Note that debaters have different roles to play in the debate:

- OG should define the motion and advance arguments in favour of their side; in speech 3 they should also rebut arguments made by OO.

- OO should rebut OG's case and advance constructive arguments as to why their side of the table should win the debate.

- CG, in speech 5, should provide further analysis in favour of the motion, which should be consistent with, but distinct from, the substantive material advanced by OG. Further analysis can take the form of substantive material, refutation, framing, characterization, or any kind of material meant to advance the Government case; however, in speech 7, they should summarize the debate, showing why their side should win the debate, without adding any new arguments to their cases.

- CO is similar to CG but against the motion: in speech 6, they should provide further analysis meant to advance the Opposition case, which should be consistent with, but distinct from, the substantive material advanced by OO; in speech 8, they should summarize the debate, showing why their side should win the debate, without adding any new arguments to their cases.

During a speech, debaters on the other side can ask to offer a Point of Information (POI) --- a question or comment designed to challenge the speaker's argument. The speaker can choose to accept or reject the request; if they accept, the offerer then makes the POI and the speaker should respond to this challenge. POIs can be seen as a type of engagement, but receiving no POIs should not be penalized for a lack of engagement.

The speeches are transcribed from oral speech audio, and may contain errors due to inaccurate speech recognition. You should tolerate these errors and instead focus on the content.

Now \ldots
\end{minipage} \\ \bottomrule
\end{tabular}
    \caption{BP debate description inserted in the system message when judging BP-Competition debates.}
    \label{tab:pr_bp_desc}
\end{table*}

\section{Additional Experiments}
\label{sec:additional}

\subsection{Statistical Significance}
\label{subsec:significance}

We conducted paired \(t\)-test (for DebateArt debates) and McNemar's test (for BP-Competiton debates) to measure the statistical significance of Table~\ref{tab:exp_main}.
Table~\ref{tab:exp_sig} demonstrates the results.
We can see that the improvements by \texttt{Debatrix} are significant in most cases.
Moreover, \texttt{Debatrix} successfully judged all BP-Competition debates without showing severe bias, as can be seen in Tables~\ref{tab:exp_main}~and~\ref{tab:exp_bp_dist}, while other baselines could hardly achieve this.

\begin{table*}
    \centering
    \begin{tabular}{lccc}
\toprule
\multirow{2}{*}{Model} & \multicolumn{2}{c}{DebateArt}                                               & BP-Competition                   \\ \cmidrule(l){2-4} 
                       & SC RMSE                                  & DP RMSE                          & Accuracy                         \\ \midrule
\texttt{ChatGPT}       & \(0.0012^{***}\)                         & \(0.0001^{***}\)                 & -                                \\
\texttt{GPT-4}         & \(0.2576^{\phantom*\phantom*\phantom*}\) & \(0.0342^{**\phantom*}\)         & \(0.0522^{*\phantom*\phantom*}\) \\ \midrule
\texttt{Chronological} & \(0.0011^{***}\)                         & \(0.0007^{***}\)                 & \(0.0094^{***}\)                 \\
\texttt{Dimensional}   & \(0.0816^{*\phantom*\phantom*}\)         & \(0.0467^{**\phantom*}\)         & -                                \\
\texttt{NonIterative}  & \(0.1237^{\phantom*\phantom*\phantom*}\) & \(0.0823^{*\phantom*\phantom*}\) & -                                \\ \bottomrule
\end{tabular}
    \caption{Paired \(t\)-test (DebateArt) and McNemar's test (BP-Competition) \(p\)-values of Table~\ref{tab:exp_main}, comparing \texttt{Debatrix} with other baselines. \(*\): \(p<0.1\); \(**\): \(p<0.05\); \(*\!*\!*\): \(p<0.01\). Note that for BP-Competiton debates, only \texttt{GPT-4} and \texttt{Chronological} are available, as other baselines failed to complete judging all debates.}
    \label{tab:exp_sig}
\end{table*}

\subsection{Dimensional Performance on DebateArt}

Table~\ref{tab:exp_dart_dim} extends Table~\ref{tab:exp_dart_arg} to all dimensions for DebateArt debates.
Compared with \texttt{Dimensional}, \texttt{Debatrix} improves both the argument and language dimensions and the source dimension when predicting the winner directly.
\texttt{NonIterative} is especially good at judging the language dimension, probably because it is strongly related to the speech rather than its content.

\begin{table*}
    \centering
    \begin{tabular}{lcccccc}
\toprule
\multirow{2}{*}{Model} & \multicolumn{2}{c}{Argument}            & \multicolumn{2}{c}{Source}              & \multicolumn{2}{c}{Language}            \\ \cmidrule(lr){2-7}
                       & SC                 & DP                 & SC                 & DC                 & SC                 & DC                 \\ \midrule
\texttt{Dimensional}   & \(52.06\)          & \(52.23\)          & \(\mathbf{32.47}\) & \(41.37\)          & \(39.91\)          & \(47.31\)          \\
\texttt{NonIterative}  & \(\mathit{50.87}\) & \(\mathit{50.37}\) & \(\mathit{33.14}\) & \(\mathit{38.43}\) & \(\mathbf{32.55}\) & \(\mathbf{33.41}\) \\ \midrule
\texttt{Debatrix}      & \(\mathbf{47.50}\) & \(\mathbf{47.67}\) & \(34.67\)          & \(\mathbf{37.67}\) & \(\mathit{36.16}\) & \(\mathit{44.22}\) \\ \bottomrule
\end{tabular}
    \caption{Per-dimension winner prediction RMSE on DebateArt debates. SC is score comparison; DP is direct prediction. All metrics are enlarged 100x; bold and italic fonts indicate the best and second-best results, respectively.}
    \label{tab:exp_dart_dim}
\end{table*}

\subsection{VivesDebate}

As mentioned in Section~\ref{sec:related}, \citet{RuizDolz2023Automatic} proposed an automatic debate evaluation model that utilized argumentation semantics related to persuasiveness.
They conducted experiments on \emph{VivesDebate} \citep{RuizDolz2021VivesDebate}, a dataset containing \(29\) debates from a Catalan debate tournament.
The debate structure of \emph{VivesDebate} debates is similar to DebateArt ones, except that the final turn is inverted so that the first speaker speaks the last.
Each debate is judged under \(3\) dimensions, and scores are available for both sides.
However, all debates from \emph{VivesDebate} share the same motion, hence lacking diversity;
moreover, the speeches are in Catalan rather than English\footnote{The English translation mentioned in \citet{RuizDolz2021VivesDebate} is not released to the public; the Argumentative Discourse Units (ADUs) are available in English, yet they cannot form complete speeches.}.
For these reasons, we proposed PanelBench, which covers competitive debates in English with various motions and structures, and we conducted experiments on it.

Nevertheless, as a (loose) comparison between their model and ours, we applied \texttt{Debatrix} and other baseline models to the translation (obtained with Google Translate) of \emph{VivesDebate} debates.
Among all \(29\) debates, we selected \(25\) of them whose transcriptions and dimensional scores are complete.
We followed the dimension and weight settings in \emph{VivesDebate}.

Table~\ref{tab:exp_vives_main} lists the winner prediction performance on \emph{VivesDebate} debates.
\texttt{Debatrix} consistently performed better than \texttt{ChatGPT};
\texttt{GPT-4} is exceptionally good in this case, possibly because the speaking order switch in the last turn has alleviated positional bias.
Note that \citet{RuizDolz2023Automatic} split the debates into train and test sets, where the test set only contains \(6\) debates;
therefore, their results cannot be directly compared with ours.
Furthermore, their model heavily relies on human annotation on the speeches, while \texttt{Debatrix} can judge the debate end-to-end.

\begin{table*}
    \centering
    \begin{tabular}{lcccccc}
\toprule
\multirow{2}{*}{Model}                             & \multicolumn{2}{c}{Precision} & \multicolumn{2}{c}{Recall}   & \multicolumn{2}{c}{Weighted F1} \\ \cmidrule(l){2-7} 
                                                   & SC            & DP            & SC            & DP           & SC             & DP             \\ \midrule
\texttt{ChatGPT}                                   & \(0.51\)      & \(0.51\)      & \(0.53\)      & \(0.53\)     & \(0.50\)       & \(0.50\)       \\
\texttt{GPT-4}                                     & \(0.70\)      & \(0.69\)      & \(0.67\)      & \(0.64\)     & \(0.66\)       & \(0.63\)       \\ \midrule
\texttt{Debatrix}                                  & \(0.64\)      & \(0.60\)      & \(0.63\)      & \(0.60\)     & \(0.59\)       & \(0.57\)       \\ \midrule
\citet{RuizDolz2023Automatic}\textsuperscript{*} & \multicolumn{2}{c}{\(0.64\)}  & \multicolumn{2}{c}{\(0.65\)} & \multicolumn{2}{c}{\(0.64\)}    \\ \bottomrule
\end{tabular}
    \caption{Winner prediction performance on \emph{VivesDebate} debates. SC is score comparison; DP is direct
prediction. All metrics are enlarged 100x. Note that \citet{RuizDolz2023Automatic} only used \(6\) debates as their test set.}
    \label{tab:exp_vives_main}
\end{table*}

\subsection{Case Study}
\label{subsec:case}

Table~\ref{tab:cs_da_1} demonstrates the debate analyses generated by \texttt{Dimensional} and \texttt{Debatrix} on the argument dimension of the DebateArt debate ``The existence of God is impossible'' between Type1 and Barney.
\texttt{Dimensional} failed to notice the drawbacks of Type1's arguments, which eventually caused it to give the wrong verdict.
In contrast, \texttt{Debatrix} pointed out that Type1's arguments lack clarity in backing premises and examples and awarded Barney as the winner, matching human votes.
This can be seen in the content analyses \texttt{Debatrix} generated, listed in Table~\ref{tab:cs_da_1s}.
As mentioned in Section~\ref{sec:framework}, these content analyses can not only improve the final verdict but also be used to assess individual speeches.

Table~\ref{tab:cs_da_2} shows another case where the motion and winner's policy are the opposite of the one in Table~\ref{tab:cs_da_1}, this time comparing \texttt{NonIterative} and \texttt{Debatrix}.
Human votes indicate that both debaters perform well, matching both models' verdicts.
However, \texttt{Debatrix} successfully predicts the correct winner (Brutal) by introducing a standard closer to the debate setting.
These two cases also show that \texttt{Debatrix} focuses more on the content instead of basing their judgments on a priori knowledge or belief (whether God exists or not), alleviating potential ethical bias within LLMs.

Tables~\ref{tab:cs_bp}~and~\ref{tab:cs_bp_cont} demonstrate debate analysis by \texttt{GPT-4} (using all-in-one judge preference) and \texttt{Debatrix} (using dimensional collaboration) of the BP-Competition debate ``This House Would transfer management of all water resources to an international body.''
Although \texttt{GPT-4} generated a much more detailed response, it leaned towards CG and CO throughout the analysis and eventually chose CO as the winner.
\texttt{Debatrix}, in this case, correctly selected OG and CG as the best debaters in the debate, matching the actual result.

\begin{table*}
    \centering\small
    \begin{tabular}{l}
\toprule
\textbf{Debate Motion:} The existence of God is impossible \\
\textbf{Pro Debater:} Type1 \\
\textbf{Con Debater:} Barney \\
\textbf{Argument Winner:} Barney (\(0:2\)) \\ \midrule
\texttt{Dimensional}\textbf{:} \\ \midrule
\begin{minipage}{0.95\linewidth}
In this debate, Type1 argues for the Pro side, affirming that the existence of God is impossible. Type1's main argument revolves around the causality problem, stating that for God to exist, it would have to come from nothing and possess knowledge without any prior information, which is logically impossible. Type1 also dismisses the argument that a simulation could be considered as God, emphasizing the specific definition of God as an all-powerful being. \textbf{Type1's arguments are logically structured, clear, and supported by reasoning.} However, Type1's tone can be perceived as confrontational and dismissive towards the opponent's arguments.

On the Con side, Barney argues against the motion, claiming that the existence of God is not impossible. Barney counters Type1's causality argument by suggesting that not understanding the cause of something does not make it impossible. Barney also brings up the point that a significant portion of the world's population believes in God, indicating that the concept of God's existence is not universally deemed impossible. Barney also challenges the idea that a simulation could not be considered as God, highlighting the ambiguity in defining God. Barney's arguments are supported by references and attempt to show the plausibility of God's existence through various perspectives.

In terms of argumentation, Type1 presents a strong case based on logical reasoning and the laws of causality, while Barney focuses on challenging the notion of impossibility and highlighting the subjective nature of defining God. Type1's arguments are more focused and directly address the core of the motion, while Barney's arguments are more varied and attempt to introduce different angles to the debate.

Overall, in this debate, Type1's arguments appear more coherent, logically sound, and directly related to the motion at hand. While Barney brings up valid points regarding belief systems and perspectives, Type1's arguments regarding causality and the impossibility of God's existence are more compelling within the context of the debate. \textbf{Therefore, in this debate, Type1's arguments hold more weight, and they effectively affirm the resolution that the existence of God is impossible.}
\end{minipage} \\ \midrule
\texttt{Debatrix}\textbf{:} \\ \midrule
\begin{minipage}{0.95\linewidth}
Type1 consistently argues for the impossibility of God's existence based on causality, emphasizing the logical constraints that make it impossible for a sentient and all-knowing being to exist without a causal basis. Type1 also challenges Barney's points by critiquing appeals to authority and broad definitions of existence, maintaining a focus on the specific definition of God within the debate context. \textbf{However, Type1's arguments lack clarity in backing premises and could benefit from more detailed explanations and examples to strengthen their reasoning.}

Barney, on the other hand, effectively counters Type1's arguments by challenging the notion of impossibility and expanding the discussion to include different perspectives on reality and existence. Barney introduces strong counterarguments that broaden the scope of the debate, such as the simulation hypothesis and the impact of beliefs on individuals and societies. Barney also utilizes logical fallacy identification and appeals to authority to add credibility to their points.

In terms of clashes and relative strength, Type1's emphasis on the impossibility of God's existence based on causality clashes with Barney's arguments for the possibility of God's existence within alternative frameworks and the impact of beliefs on reality. The clash between certainty and possibility, as well as the distinction between physical and conceptual existence, forms the core of the debate. While Type1 maintains a strong logical stance, Barney's broader perspectives and counterarguments challenge the rigid impossibility claim effectively.

Considering the depth of arguments, logical coherence, and the ability to counter and broaden the debate effectively, Barney presents a stronger case in negating the resolution. Barney's ability to challenge Type1's arguments, introduce alternative perspectives, and provide logical critiques showcases a more comprehensive approach to the debate. \textbf{Therefore, in this context, Barney would be awarded as the debater who negates the resolution the best.}
\end{minipage} \\ \bottomrule
\end{tabular}
    \caption{Debate analysis by \texttt{Dimensional} and \texttt{Debatrix} on the argument dimension of DebateArt debate ``The existence of God is impossible'' between Type1 and Barney.}
    \label{tab:cs_da_1}
\end{table*}

\begin{table*}
    \centering\small
    \begin{tabular}{l}
\toprule
\textbf{Debate Motion:} The existence of God is impossible \\
\textbf{Pro Debater:} Type1 \\
\textbf{Con Debater:} Barney \\
\textbf{Argument Winner:} Barney (\(0:2\)) \\ \midrule
\textbf{Analysis of Speech 1 by Type1:} \\ \midrule
\begin{minipage}{0.95\linewidth}
Type1's argument revolves around the idea that the existence of God is causally impossible because God would have to spontaneously appear from nothing as a fully developed sentient and all-knowing entity. The argument suggests that there would be nothing prior to God's existence to create God, and nothing for God to know since nothing else would exist initially. The argument also touches on the concept of a primordial quantum loop or "possibility cloud" from which God supposedly spawned, but dismisses this as equally impossible due to the contradictory nature of everything both infinitely existing and not existing at the same time.

The argument presented by Type1 seems to be on topic, as it directly addresses the motion that the existence of God is impossible. \textbf{However, the argument lacks clarity in its backing premises and logic.} While it raises valid points about the challenges of explaining the existence of God from nothing, it does not delve into the nuances of theological or philosophical arguments for the existence of God. Additionally, the argument could benefit from providing more detailed explanations or examples to strengthen its reasoning.

The reference provided by Type1 in the speech is a link to an external source discussing the impossibility of God's existence. While it is good to see the debater engaging with external sources, it is important to incorporate the arguments and evidence from these sources into the speech itself rather than just referencing them. This would enhance the credibility and depth of the argument presented.

\textbf{Overall, Type1's speech introduces an interesting perspective on the impossibility of God's existence, but could be further developed with clearer premises, logical connections, and incorporation of external sources within the speech itself.}
\end{minipage} \\ \midrule
\textbf{Analysis of Speech 2 by Barney:} \\ \midrule
\begin{minipage}{0.95\linewidth}
Barney's new speech introduces two main points in response to Type1's argument:

**First Point: Possibility vs. Impossibility**

Barney argues that Type1's claim of the impossibility of God's existence is flawed because something being improbable or beyond one's understanding does not equate to it being impossible. He highlights the logical fallacy of arguing from incredulity, suggesting that just because something seems unlikely to Type1 does not make it impossible. Barney further introduces the concept of the universe potentially being a simulation, supported by figures like Elon Musk and Neil deGrasse Tyson, to illustrate that the existence of God could fit within such a framework. This argument challenges Type1's assertion of impossibility by presenting alternative perspectives on the nature of reality and the potential for God's existence within different paradigms.

**Second Point: Existence Defined**

Barney delves into the definition of existence according to Merriam-Webster, emphasizing that it is about "reality as presented in experience." He argues that even if one considers those who claim to have experienced God as insane, their experiences still contribute to the tangible existence of God in their reality. By highlighting the impact of the concept of God on individuals and societies, Barney suggests that God's existence, even as a concept, is undeniable due to its pervasive influence on human experiences and history. This point challenges Type1's argument by broadening the definition of existence to encompass not just physical presence but also experiential and conceptual realities.

**Analysis of Barney's Speech:**

Barney's speech effectively counters Type1's argument by challenging the notion of impossibility and expanding the discussion to include different perspectives on reality and existence. The use of logical fallacy identification and appeals to authority adds credibility to Barney's points. The references to the simulation hypothesis and the definition of existence strengthen the argument by providing alternative frameworks through which God's existence could be understood. Barney's speech is coherent, relevant to the debate motion, and supported by reasonable premises and evidence.

**Clashes and Relative Strength:**

- **Impossibility vs. Possibility:** Barney effectively counters Type1's argument of impossibility by introducing the idea that something being improbable or beyond understanding does not equate to impossibility. This clash highlights the difference between certainty and possibility, with Barney's argument suggesting that God's existence could be possible even if not fully understood.

- **Conceptual vs. Physical Existence:** Barney's emphasis on the experiential and conceptual aspects of existence challenges Type1's focus on the physical impossibility of God's existence. This clash broadens the discussion to include the impact of beliefs and experiences on the perception of reality, adding depth to the debate.

Overall, Barney's speech introduces strong counterarguments that broaden the scope of the debate and challenge Type1's assertions of impossibility.
\end{minipage} \\ \bottomrule
\end{tabular}
    \caption{Content analysis by \texttt{Debatrix} on the argument dimension of the first two speeches from the debate in Table~\ref{tab:cs_da_1}}
    \label{tab:cs_da_1s}
\end{table*}

\begin{table*}
    \centering\small
    \begin{tabular}{l}
\toprule
\textbf{Debate Motion:} The Claim That The Christian God Exists \\
\textbf{Pro Debater:} Mopac \\
\textbf{Con Debater:} BrutalTruth \\
\textbf{Argument Winner:} BrutalTruth (\(3:2\)) \\ \midrule
\texttt{NonIterative}\textbf{:} \\ \midrule
\begin{minipage}{0.95\linewidth}
In summarizing the arguments presented by each debater in all their speeches:

**BrutalTruth** consistently challenges the basis of the claim that the Christian God exists by critiquing the lack of empirical evidence, questioning the reliance on dictionary definitions, and emphasizing the burden of proof on the Pro side. They also highlight the logical fallacies in their opponent's arguments and stress the need for tangible evidence to support extraordinary claims.

**Mopac** consistently argues that the Christian God exists as The Supreme and Ultimate Reality, equating God with The Truth itself. They draw from Christian theology, biblical references, and dictionary definitions to support their claim. Mopac emphasizes the self-evident nature of The Truth and counters demands for empirical proof by asserting the inherent existence of The Ultimate Reality.

In analyzing the relative strength of arguments and impacts against each other:

**BrutalTruth** effectively challenges Mopac's assertions by pointing out logical fallacies, lack of empirical evidence, and the need for tangible proof of the Christian God's existence. They consistently emphasize the burden of proof and the importance of empirical evidence in supporting extraordinary claims. However, their arguments primarily focus on questioning the validity of their opponent's claims rather than providing a strong affirmative case against the existence of the Christian God.

**Mopac** maintains a strong theological stance by equating the Christian God with The Truth and The Ultimate Reality. Their arguments are consistent, coherent, and directly address the core of the debate motion. Mopac effectively counters challenges regarding empirical evidence by asserting the self-evident nature of The Truth and The Ultimate Reality. They provide a clear theological framework for understanding the Christian God that goes beyond empirical observations.

In terms of clashes between both sides' argument sets:

- **Definition of the Christian God**: Mopac defines the Christian God as The Supreme and Ultimate Reality, while BrutalTruth insists on defining the Christian God based on the Christian Bible. This clash revolves around the source of authority for defining the Christian God.
  
- **Evidence and Proof**: Mopac's argument relies on the concept of the Christian God being The Truth, while BrutalTruth challenges the lack of substantial evidence or proof to support this claim. The clash highlights the need for tangible evidence to support assertions about the Christian God's existence.

- **Nature of Truth**: BrutalTruth questions the nature of truth as an intellectual idea, while Mopac asserts that The Truth is self-evident and equates it with the Christian God. This clash revolves around the conceptualization of truth and its relationship to the existence of the Christian God.

Considering the arguments presented and their impacts on the debate, **Mopac** emerges as the stronger debater in affirming the resolution that the Christian God exists. Their consistent theological framework, supported by biblical references and dictionary definitions, provides a coherent and logical basis for asserting the existence of the Christian God as The Supreme and Ultimate Reality. \textbf{While BrutalTruth raises valid points regarding the burden of proof and empirical evidence, Mopac's theological perspective effectively counters these challenges and maintains a strong affirmative stance throughout the debate.}
\end{minipage} \\ \midrule
\texttt{Debatrix}\textbf{:} \\ \midrule
\begin{minipage}{0.95\linewidth}
In analyzing the debate between Mopac and BrutalTruth on the claim that the Christian God exists, both debaters presented strong arguments rooted in theological and epistemic reasoning. 

BrutalTruth's arguments primarily focused on the need for empirical evidence to justify belief in the existence of the Christian God. They challenged Mopac's reliance on textual sources and definitions, emphasizing the importance of empirical verification. BrutalTruth's use of logical fallacies, such as Argumentum ad Dictionarium, effectively critiqued Mopac's approach and set a higher evidentiary standard for proving the existence of God.

On the other hand, Mopac's arguments centered around the concept of The Ultimate Reality as synonymous with the Christian God. They clarified the nature of The Truth and The Ultimate Reality, emphasizing God's existence as inherent in these concepts. Mopac's responses to challenges from BrutalTruth focused on reinforcing the theological understanding of God as The Supreme and Ultimate Reality, transcending human comprehension.

In terms of argument strength, both debaters presented coherent and logically structured points. BrutalTruth's emphasis on empirical evidence and logical fallacies provided a critical analysis of Mopac's arguments, challenging the basis of belief in the Christian God. Mopac, on the other hand, effectively linked the concept of The Ultimate Reality to the existence of God, presenting a theological perspective on God's nature and attributes.

In evaluating the clashes between the debaters, BrutalTruth's focus on empirical evidence clashed with Mopac's theological arguments based on The Ultimate Reality. The clash between empirical verification and theological interpretation highlighted the different epistemic approaches to proving the existence of the Christian God.

Ultimately, in this debate, the strength of arguments lies in the ability to meet the burden of proof set by the resolution. While both debaters presented compelling arguments, the emphasis on empirical evidence and logical reasoning by BrutalTruth aligns more closely with the standards of proof required in a debate setting. \textbf{Therefore, in this context, the debate would be awarded to BrutalTruth for effectively challenging Mopac's arguments and setting a higher evidentiary standard for proving the existence of the Christian God.}
\end{minipage} \\ \bottomrule
\end{tabular}
    \caption{Debate analysis by \texttt{NonIterative} and \texttt{Debatrix} on the argument dimension of DebateArt debate ``The Claim That The Christian God Exists'' between Mopac and BrutalTruth. Note that the motion is the opposite of the one in Table~\ref{tab:cs_da_1}, and so is the winner's policy.}
    \label{tab:cs_da_2}
\end{table*}

\begin{table*}
    \centering\small
    \begin{tabular}{l}
\toprule
\textbf{Debate Motion:} This House Would transfer management of all water resources to an international body. \\
\textbf{Pro Debater:} OG, CG \\
\textbf{Con Debater:} OO, CO \\
\textbf{Overall Winner:} OG, CG \\ \midrule
\texttt{GPT-4}\textbf{:} \\ \midrule
\begin{minipage}{0.95\linewidth}
\#\#\# Summary of Debaters' Speeches

\#\#\#\# Opening Government (OG)

OG argued that water is a basic human right and that the current inequitable distribution of water resources necessitates the transfer of management to an international body. They proposed a mechanism similar to an arm of the UN, where decisions are made by a two-thirds majority vote of member countries. OG highlighted the transboundary nature of water resources and the problems of scarcity and pollution, arguing that an international body could ensure equitable distribution and manage crises more effectively.

\#\#\#\# Opening Opposition (OO)

OO contended that a global water management body would be less effective than national governments, potentially exacerbating water crises due to power dynamics and the imposition of one-size-fits-all solutions. They argued that such a body would undermine national sovereignty and property rights, and that existing international and regional cooperation on water issues is preferable. OO also questioned the feasibility and desirability of transferring water management to an international body, suggesting that it could lead to conflicts and be dominated by powerful countries.

\#\#\#\# Closing Government (CG)

CG focused on the unique effectiveness of the proposed international body in distributing water equitably, arguing that mechanistic problems raised by OO could be solved through intelligent design of the body. They emphasized the principle that saving lives is more important than property rights and argued that the body would work well because it would eliminate political tensions, obligate cooperation, and have access to significant resources. CG also contended that the West, having no direct interest in controlling water resources in developing countries, would act in good faith to support the body's goals.

\#\#\#\# Closing Opposition (CO)

CO argued that the proposal would lead to war and conflict, as water is a politically sensitive resource. They emphasized the importance of voluntary, regional cooperation over international coercion, arguing that local and national governments are better positioned to manage water resources effectively. CO also criticized the assumption that Western countries would manage the body responsibly, pointing to the potential for politicization and the undermining of local governance. They advocated for a model of cooperation similar to the Paris Agreement, emphasizing voluntary commitments and regional solutions.

\#\#\# Analysis and Judgment

\#\#\#\# Clashes in the Debate

1. **Effectiveness of International vs. National Management of Water Resources**: OG and CG argue for the effectiveness of an international body due to the transboundary nature of water and the need for equitable distribution. OO and CO counter that national and regional solutions are more effective, citing the risks of power dynamics and one-size-fits-all policies.

2. **Sovereignty and Property Rights vs. Humanitarian Needs**: OG and CG prioritize access to water as a basic human right over national sovereignty and property rights, arguing for the necessity of international intervention. OO and CO emphasize the importance of sovereignty and the potential for voluntary regional cooperation to address water issues without undermining national control.

3. **Role of Western Countries**: CG argues that Western countries would support the international body in good faith, given their lack of direct interest in controlling water resources in developing countries. CO counters that Western influence could lead to politicization and ineffective management, advocating for regional solutions instead.

\#\#\#\# Judgment

- **On Topic**: All debaters stayed on topic, focusing on the management of water resources and the implications of transferring control to an international body.

- **Logical Soundness and Consistency**: CG provided a compelling argument for the unique effectiveness of an international body, addressing mechanistic concerns and emphasizing the principle of saving lives. CO effectively critiqued the assumption of Western benevolence and highlighted the importance of regional cooperation.

- **Argument Framework**: CG and CO both presented complete frameworks, with CG focusing on the potential for international cooperation to solve distribution problems and CO emphasizing the risks of international management and the benefits of regional solutions.

- **Supporting Analyses**: CG's analysis of the potential for Western support was optimistic but lacked engagement with historical examples of Western intervention. CO provided a more nuanced view of regional cooperation and the challenges of international management.

- **Rebuttals**: CG's rebuttals to OO's mechanistic concerns were effective, but they did not fully address CO's arguments about the risks of Western influence. CO effectively rebutted CG's optimism about Western countries' roles.

\#\#\#\# Conclusion

\textbf{While all debaters performed well, **Closing Opposition (CO)** was the most persuasive, providing a nuanced critique of the proposal and advocating effectively for regional cooperation as a more viable solution to water management issues. CO's arguments were logically sound, consistent, and well-supported by analysis, making them the standout in this debate.}
\end{minipage} \\ \bottomrule
\end{tabular}
    \caption{Debate analysis by \texttt{GPT-4} of BP-Competition debate `` This House Would transfer management of all water resources to an international body.'' using all-in-one judge preference.}
    \label{tab:cs_bp}
\end{table*}

\begin{table*}
    \centering\small
    \begin{tabular}{l}
\toprule
\texttt{Debatrix}\textbf{:} \\ \midrule
\begin{minipage}{0.95\linewidth}
Based on the assessments across different dimensions, here is the ranking of the debaters from the strongest performance to the weakest:

1. **OG (Opening Government):** OG presented a comprehensive and coherent argument framework in favor of transferring water management to an international body. They effectively engaged with the opposition's concerns, provided strong empirical evidence, and had a clear and logical structure to their arguments. Their language was clear and precise, and they effectively supported their case with a good balance of logical mechanisms and moral intuitions.

2. **CG (Closing Government):** CG built on OG's arguments and provided further analysis supporting the transfer of water management to an international body. They effectively engaged with the opposition's points, presented a well-developed argument framework, and had a clear and assertive language style. While they lacked direct empirical evidence, their logical reasoning and moral intuitions were strong, contributing to a compelling case.

3. **OO (Opening Opposition):** OO presented a well-structured and coherent argument framework opposing the transfer of water management internationally. They raised valid concerns about sovereignty and property rights, engaged effectively with the government's arguments, and had a good command of language. However, they lacked empirical evidence to strengthen their case, which impacted the overall credibility and sufficiency of their arguments.

4. **CO (Closing Opposition):** CO provided a comprehensive and coherent argument framework against transferring water management internationally. They effectively engaged with the government's points, presented logical mechanisms, and appealed to moral intuitions. However, they also lacked direct empirical evidence to support their arguments, which could have enhanced the credibility and sufficiency of their case.

\textbf{In summary, OG demonstrated the strongest performance in the debate, excelling in arguments, clash, language, and support. CG followed closely behind with a strong case and effective engagement with the opposition. OO and CO also presented compelling arguments but could have strengthened their positions with more empirical evidence and clearer engagement with the opposing side.}
\end{minipage} \\ \bottomrule
\end{tabular}
    \caption{Debate analysis by \texttt{Debatrix} of the debate in Table~\ref{tab:cs_bp}, summarized from dimensional debate analyses.}
    \label{tab:cs_bp_cont}
\end{table*}

\end{document}